%% file: main.tex
\documentclass[10pt,twocolumn,letterpaper]{article}

\usepackage{cvpr}
\usepackage{times}
\usepackage{epsfig}
\usepackage{graphicx}
\usepackage{amsmath}
\usepackage{amssymb}
\usepackage{diagbox}
\usepackage{graphicx}
\usepackage{comment}
\usepackage{amsmath,amssymb}
\usepackage{color}
\usepackage{pifont}
\usepackage{subcaption}

\usepackage{soul} 
\usepackage[dvipsnames]{xcolor}
\usepackage{booktabs}
\usepackage{multirow}
\usepackage{caption}
\usepackage{wrapfig}

\usepackage{paralist}
\usepackage{microtype}
\usepackage{arydshln}
\usepackage{chngcntr}
\usepackage{url}
\usepackage{makecell}
\usepackage{authblk}

\usepackage{gensymb}
\usepackage{enumitem}

\usepackage[pagebackref=true,breaklinks=true,letterpaper=true,colorlinks,bookmarks=false]{hyperref}

\newcommand{\cmark}{\text{\ding{51}}}
\newcounter{magicrownumbers}
\newcommand\rownumber{\stepcounter{magicrownumbers}\arabic{magicrownumbers}}

\usepackage[pagebackref=true,breaklinks=true,letterpaper=true,colorlinks,bookmarks=false]{hyperref}

\cvprfinalcopy 

\def\cvprPaperID{10173}

\ifcvprfinal\fi

\begin{document}
\ifcvprfinal\pagestyle{empty}\fi
\title{TextOCR: Towards large-scale end-to-end reasoning \\ for arbitrary-shaped scene text}

\author{Amanpreet Singh\thanks{Equal Contribution. Correspondence to \href{mailto:textvqa@fb.com}{textvqa@fb.com}} , Guan Pang$^\ast$, Mandy Toh$^\ast$, \\Jing Huang, Wojciech Galuba, and Tal Hassner}

\affil{Facebook AI Research\\\href{https://textvqa.org/textocr}{https://textvqa.org/textocr}}

\newcommand{\datasetName}{TextOCR\xspace}

\maketitle
\thispagestyle{plain}
\ifcvprfinal\pagestyle{plain}\fi

\begin{figure*}[h]
    \centering
    \includegraphics[width=0.9\textwidth]{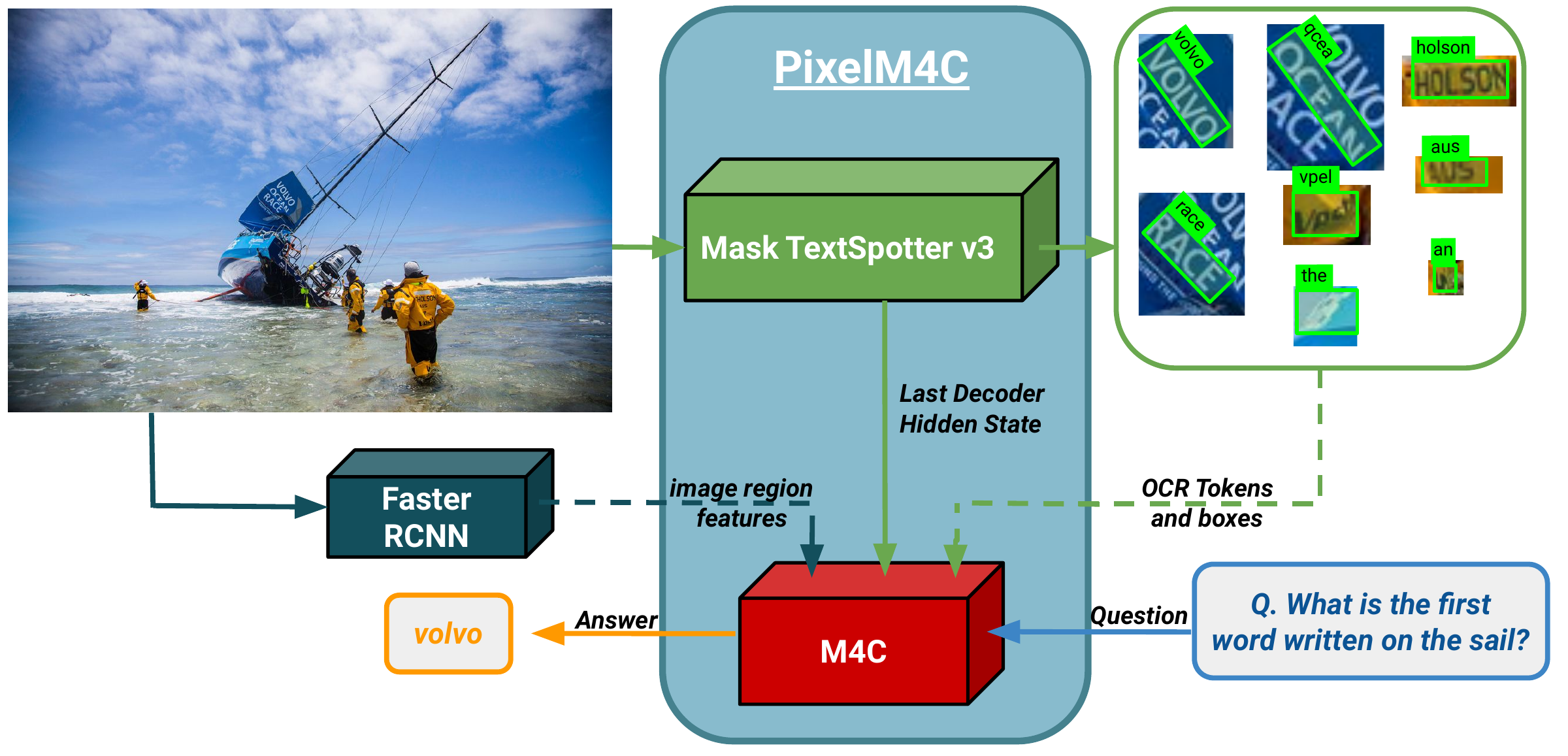}\vspace{-1mm}
    \captionof{figure}{\textbf{PixelM4C - An end-to-end TextVQA model.} In this work, we bridge the gap between arbitrary scene-text detection/recognition and scene-text based reasoning in TextVQA~\cite{singh2019towards} and TextCaps~\cite{sidorov2020textcaps}. We introduce \datasetName, largest real scene-text detection and recognition dataset with 900k annotated arbitrary-shaped words collected on TextVQA images. Further, we build PixelM4C, an end-to-end TextVQA model which uses \datasetName trained Mask TextSpotter v3~\cite{liao2020maskv3} and M4C~\cite{hu2020iterative} models to do text-based reasoning directly on the images unlike previous works which rely on pre-extracted OCR text and features. The solid lines in the figure show backpropagable paths.
    }\vspace{-3mm}
    \label{fig:pipeline}
\end{figure*}

\begin{abstract}
A crucial component for the scene text based reasoning required for TextVQA and TextCaps datasets involve detecting and recognizing text present in the images using an optical character recognition (OCR) system. The current systems are crippled by the unavailability of ground truth text annotations for these datasets as well as lack of scene text detection and recognition datasets on real images disallowing the progress in the field of OCR and evaluation of scene text based reasoning in isolation from OCR systems. In this work, we propose \datasetName, an arbitrary-shaped scene text detection and recognition with 900k annotated words collected on real images from TextVQA dataset. We show that current state-of-the-art text-recognition (OCR) models fail to perform well on \datasetName and that training on \datasetName helps achieve state-of-the-art performance on multiple other OCR datasets as well. We use a \datasetName trained OCR model to create PixelM4C model which can do scene text based reasoning on an image in an end-to-end fashion, allowing us to revisit several design choices to achieve new state-of-the-art performance on TextVQA dataset.
\end{abstract}

\section{Introduction}
The computer vision community has recently seen a surge in interest to understand and reason on the text present in the images (scene text) beyond the OCR extraction. In particular, multiple datasets have been introduced that focus on \textbf{visual question answering (VQA)} \cite{singh2019towards, biten2019scene, mishra2019ocr} and \textbf{image captioning} \cite{sidorov2020textcaps} but in the context of scene text. These tasks involve understanding the objects and text in the image and then reasoning over the spatial and semantic relations between these along with a textual input (\eg question). Though the OCR systems have matured, they still don't work well on pictures involving real-life scenarios given the lack of large annotated real scene text OCR datasets. The text extracted by the OCR systems doesn't mean anything in itself until it is used to solve a task which involves using the scene text. Other than VQA and image captioning, the potential use cases include several impactful and interesting tasks from hate speech and misinformation detection \cite{kiela2020hateful} to the study of cultural heritage~\cite{hassner2012computation,hassner2014digital}. 

Although, the field has witnessed success and progress in datasets on downstream OCR applications, the performance of state-of-the-art models on these datasets are nowhere close to human accuracy due to multiple factors which includes the quality of the OCR extracted from existing OCR systems, unavailability of ground-truth text annotations for the real-world images, and no feedback to OCR system to improve detection or extraction based on the errors in the downstream application \ie no end-to-end training.

In this paper, we introduce a new dataset, \datasetName, which aims to bridge these gaps by providing (i) high quality and large quantity text annotations on TextVQA images (ii) allowing end-to-end training of downstream application models with OCR systems and thus allowing fine-tuning of OCR pipeline based on the task involved. Prior to \datasetName, many OCR datasets exist~\cite{mishra2012iiit5k, wang2011svt, lucas2003icdar, karatzas2013icdar, karatzas2015icdar, phan2013svtp, risnumawan2014cute, chng2019tt, yao2012msra, liu2017ctw, veit2016coco, nayef2017mlt, shi2017rctw, nayef2019mlt, chng2019art, sun2019lsvt} that propelled the field's development, but many of these are either relatively small, or focus mostly on outdoor or store-front scenes. As a result, OCR models trained on these datasets usually don't perform well on downstream tasks from other scene types. Moreover, existing datasets usually have a low number of words per image, making them less dense, diverse and ideal to train OCR models for tasks commonly having a high text density. As a solution, we present the \datasetName dataset that contains more than 28k images and 903k words in total, averaging 32 words per image. Jointly with existing TextVQA~\cite{singh2019towards} and TextCaps~\cite{sidorov2020textcaps} datasets, it can also serve as an OCR upper bound for researchers working on them to evaluate their methods' reasoning capabilities on a fair ground.

In addition to \datasetName, we present a novel architecture, PixelM4C, that connects an OCR model, Mask TextSpotter (MTS) v3 \cite{liao2020maskv3} with downstream TextVQA model, M4C \cite{hu2020iterative}, in an end-to-end trainable fashion, as illustrated in Figure~\ref{fig:pipeline}. Through extensive analysis and ablations possible with end-to-end PixelM4C, we revisit and improve design choices from prior work to achieve new state-of-the-art performance on TextVQA dataset \cite{singh2019towards} under comparable settings and show \datasetName's performance impact on new TextCaps dataset \cite{sidorov2020textcaps}.

In summary, our main contributions include:
\begin{itemize}[nosep,itemsep=1pt,leftmargin=1.5em,labelwidth=*,align=left]
\item A large and diverse OCR dataset with $\sim$1M arbitrary-shaped word annotations (3x larger than existing datasets), with high density of $\sim$32 words per image. 
\item Extensive experiments to evaluate \datasetName showing that it is effective both as (i) a training dataset to push OCR state-of-the-art on multiple datasets and (ii) a testing dataset to offer a new challenge to the community.
\item A new end-to-end novel architecture, PixelM4C for TextVQA and TextCaps, which connects Mask TextSpotter (MTS) v3 \cite{liao2020maskv3} to M4C \cite{hu2020iterative} allowing extensive analysis and revisiting prior work's design decisions. 
\item State-of-the-art on TextVQA \cite{singh2019towards} using OCR tokens generated from \datasetName trained OCR models and insights from PixelM4C ablations under comparable settings.
\end{itemize}

\begin{figure*}[ht]
\begin{center}
\includegraphics[width=0.83\textwidth]{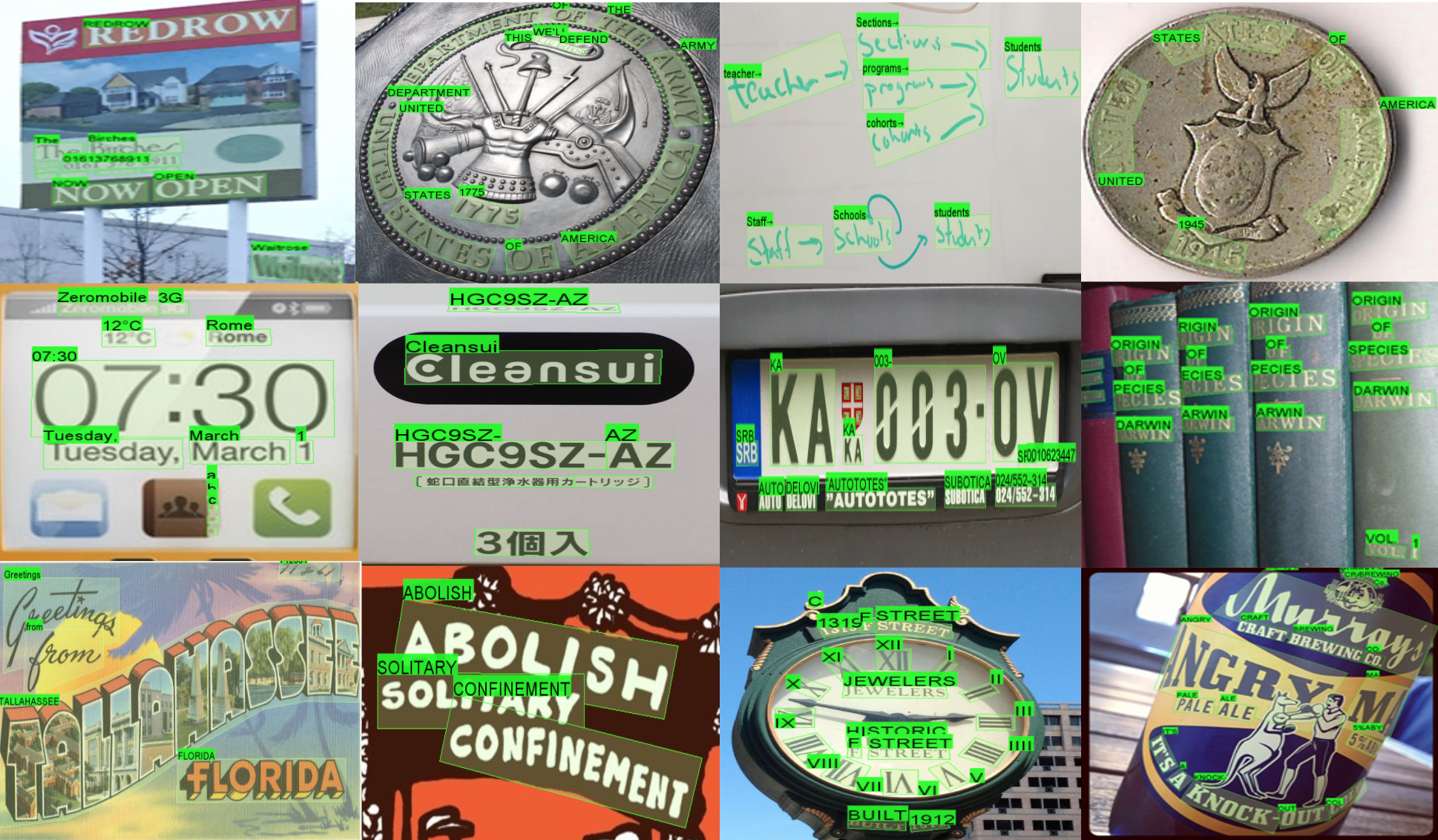}\vspace{-3mm}
\end{center}
   \caption[]{\textbf{TextOCR visualizations.} The figure shows diversity and density in \datasetName images.\footnotemark}\vspace{-3mm}
\label{fig:dataset_vis}
\end{figure*}

\section{Related work}
\subsection{OCR datasets}

\textbf{Recognition.} The text recognition datasets which involve recognizing text from cropped words can be categorized as regular or irregular. The regular datasets like IIIT5K-Words (IIIT)~\cite{mishra2012iiit5k}, Street View Text (SVT)~\cite{wang2011svt}, ICDAR2003 (IC03)~\cite{lucas2003icdar}, ICDAR2013 (IC13)~\cite{karatzas2013icdar} have horizontally aligned words while irregular datasets like ICDAR2015 (IC15)~\cite{karatzas2015icdar}, SVT Perspective (SVTP)~\cite{phan2013svtp}, CUTE80~\cite{risnumawan2014cute}, and Total Text (TT)~\cite{chng2019tt} are more challenging as they involve various transformations, such as arbitrary-oriented or curved.

\textbf{Detection.} Compared to older OCR datasets, which only allowed recognition as they came with pre-cropped words, the newer datasets can be used for either detection or end-to-end task as they have full images with labeled instances. The examples IC13\cite{karatzas2013icdar}, IC15\cite{karatzas2015icdar}, and TT\cite{chng2019tt} can use different word location formats, horizontal box, quadrilateral box, and curved polygon respectively. Additionally, datasets like MSRA-TD500~\cite{yao2012msra} and CTW1500~\cite{liu2017ctw} with line-level labels are commonly only used for detection task.

\textbf{Multi-Lingual.} In recent years, there has been a surge in large-scale multi-lingual datasets containing (i) up to 7 or 8 different scripts (\eg ICDAR17-MLT~\cite{nayef2017mlt} and ICDAR19-MLT~\cite{nayef2017mlt}), (ii) Chinese and English due to large character set (\eg ICDAR17-RCTW~\cite{shi2017rctw}, ICDAR19-ArT~\cite{chng2019art} and ICDAR19-LSVT~\cite{sun2019lsvt}). These usually use test set for a challenge not releasing the labels and being multi-lingual, the amount of data distributed in each language is smaller.

\textbf{Synthetic.} Synth90k~\cite{jaderberg2014synth90k} with 9M word-level crops, and SynthText~\cite{gupta2016synthtext} with 800K images (6M words) are the most common in OCR research. As these are synthetic and contain a lot of data, they are typically used for model pretraining or joint training with real datasets.

\footnotetext{All images are licensed under CC BY 2.0. See appendix for full attributions.}

\subsection{Downstream OCR applications}
In recent years, multiple datasets have been introduced for scene-text applications to study and reason about the text present in an image \wrt the objects in it. TextVQA \cite{singh2019towards} contains 28K images from OpenImages \cite{kuznetsova2018open} with 45K questions, each with 10 human-annotated answers, which require reading and reasoning over scene-text to answer them. Similarly, ST-VQA \cite{biten2019scene} contains 32k questions on images from 6 different sources (IC13 \cite{karatzas2013icdar}, IC15 \cite{karatzas2015icdar},
ImageNet \cite{imagenet_cvpr09}, VizWiz \cite{bigham2010vizwiz}, IIIT Scene Text Retrieval,
Visual Genome \cite{krishna2017visual}, and COCO-Text \cite{veit2016coco}). A series of datasets were introduced following these which focused on specific aspects of text-based VQA including OCR-VQA \cite{mishra2019ocr}, STE-VQA \cite{Wang_2020_CVPR}, DocVQA \cite{mathew2020docvqa}, PlotQA \cite{methani2020plotqa}, and LEAF-QA \cite{chaudhry2020leaf}.

TextCaps dataset \cite{sidorov2020textcaps} requires reading comprehension with images and contains 143K captions on TextVQA images \cite{singh2019towards}. TextCaps requires understanding how OCR words interact with each other and objects to build a caption which is coherent while tackling challenges like parts-of-speech, OCR and fixed vocabulary switching. VizWiz-Captions \cite{gurari2020captioning} also contains similar captions on VizWiz images \cite{bigham2010vizwiz} but doesn't explicitly require scene-text reasoning. 

\subsection{Downstream application models}
The state-of-the-art on TextVQA and TextCaps use the pre-extracted text tokens from a standard OCR system as additional input \cite{joulin2016fasttext}. As the OCR text can be any string, for word embeddings, we use a system that allows out-of-vocabulary words via character-level modeling or piece-based modeling \cite{joulin2016fasttext}. The other textual input (\eg question) is encoded using a pretrained word-embedding (BERT, GloVe \cite{devlin2018bert,pennington2014glove}) and fused with image's object features and OCR embeddings. The joint embedding passes through a classifier or decoder to generate the output. The state-of-the-art TextVQA model, M4C \cite{hu2020iterative}, uses transformers \cite{vaswani2017attention} to model the fusion via self and inter-modality attention to achieve 40\% on TextVQA compared to 86\% human accuracy. On TextCaps, M4C can be adapted to generate a sentence by taking previously generated words as text inputs at each time step. Multiple models have been introduced recently which ablate various components of M4C for better accuracy \cite{Kant2020SpatiallyAM,Gao2020StructuredMA,Han2020FindingTE,Jin2020RUArtAN}. Contrary to M4C and derivative works which treat OCR as a black box, in PixelM4C, we train an end-to-end model and use this capability to apply new design choices in a more informed way. To test our hypothesis, we build and compare PixelM4C with M4C as our base because of its simplicity and modular design.

\section{\datasetName dataset}
\subsection{Annotation Details}

For collecting arbitrary shaped scene text, all words within an image are annotated with polygon annotation for detection. For recognition, only Latin words are annotated. All non-Latin and illegible words are then annotated with a ``.''. Similar to COCOText, a word is defined as an uninterrupted sequence of letters. 

The annotations are performed by a set of annotators familiar with polygon word annotation. We provided the team with annotation guidelines, and a quality control training and performed rigorous checks on their work. The annotators first annotate the bounding box around a word. If the word is near-horizontal, annotators are encouraged to draw a rectangle whenever possible. If the word is curved, then the annotators draw polygon as annotations with multiple points while preserving reading direction from the first annotated point to the second annotated point. The annotators are also encouraged to annotate the bounding box with as little background space as possible. To ensure accuracy of predictions, our annotation pipeline includes an audit procedure, where expert annotators (authors) provide feedback to individual annotators for re-annotation. Please see appendix for more details on our annotation user interface.

\subsection{Statistics and Visualizations}
\begin{figure}[h]
\begin{center}
\begin{subfigure}[b]{0.65\linewidth}
\includegraphics[width=\linewidth]{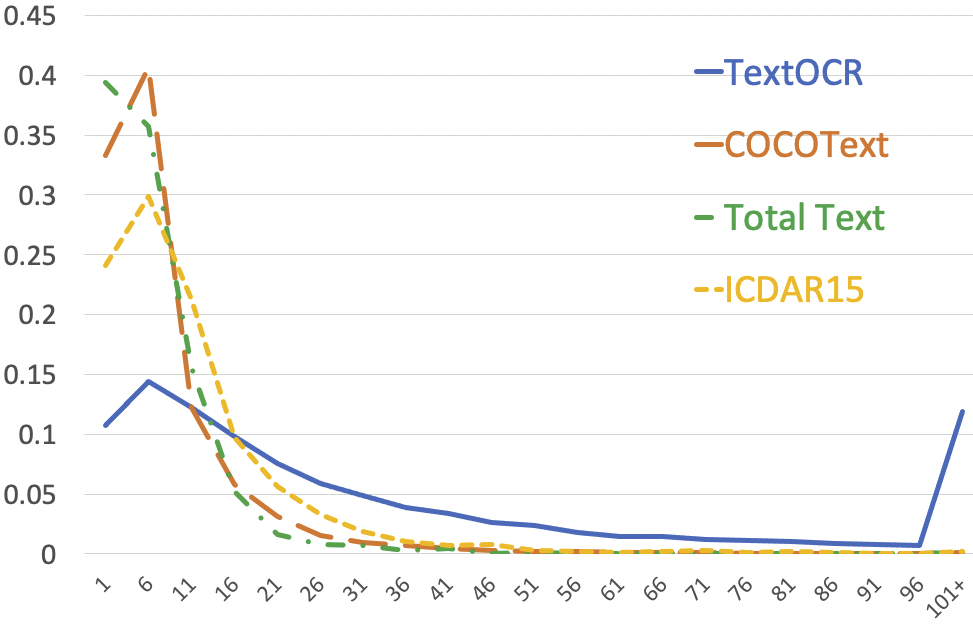}
\caption{Distribution of words per image}\label{fig:1b}
\end{subfigure}
\:
\begin{subfigure}[b]{0.3\linewidth}
\begin{subfigure}[b]{\linewidth}
\includegraphics[width=\linewidth]{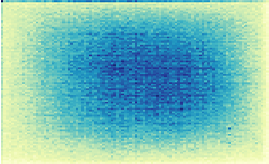}
\caption{\datasetName}\label{fig:1b}
\end{subfigure}
\begin{subfigure}[b]{\linewidth}
\includegraphics[width=\linewidth]{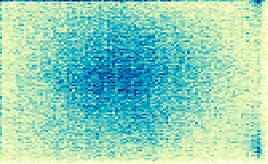}
\caption{COCOText}\label{fig:1b}
\end{subfigure}
\end{subfigure}

\end{center}
   \caption{\textbf{\datasetName distributions.} (left) Comparison of words per image showing \datasetName's higher text density, with $>$ 10\% images containing 100+ instances. (right) Word locations' heatmaps across images with blue indicating higher density. \datasetName is more uniformly annotated/distributed than COCOText \cite{veit2016coco} which is more uniform than IC13 and IC15.}
\label{fig:heatmap}
\end{figure}

Figure~\ref{fig:heatmap} shows that \datasetName is diverse both in terms of words per image (left) as well as the word locations (right). Figure~\ref{fig:heatmap} (a) compares and shows high density of word annotations in \datasetName compared with COCOText \cite{veit2016coco} and ICDAR\-15 \cite{karatzas2015icdar}. Figure~\ref{fig:heatmap} (b) and (c) compare the density of word bounding boxes in \datasetName and COCOText depicting more uniform, regular and heavy density in \datasetName suggesting that \datasetName is more precisely, uniformly and carefully annotated.

\begin{table}[t!]
\footnotesize
\setcounter{magicrownumbers}{0}
\setlength{\tabcolsep}{3pt}
\renewcommand{\arraystretch}{1}
\centering
    \begin{tabular}{@{}llrrrrr@{}}
        \toprule
        \multirow{2}[3]{*}{\bf \#} & \multirow{2}[3]{*}{\bf Dataset} & \multicolumn{2}{c}{\bf $\#$ Images} & \multicolumn{2}{c}{\bf $\#$ Words} & \multirow{2}[3]{*}{\bf \shortstack{Words \\ per Image}} \\
        \cmidrule(lr){3-4} \cmidrule(lr){5-6}
        & &  \bf Train & \bf Test & \bf Train & \bf Test & \\
        \cmidrule[\heavyrulewidth]{1-7}
        \rownumber & Synth90k~\cite{jaderberg2014synth90k}$^\dagger$ & -- & -- & 8.9M & -- & -- \\
        \rownumber & SynthText~\cite{gupta2016synthtext}$^\dagger$ & 800k & -- & 5.5M & -- & 6.9 \\
        \cmidrule{1-7}
        \rownumber & IIIT5K~\cite{mishra2012iiit5k} & -- & -- & 2000 & 3000 & -- \\
        \rownumber & SVT~\cite{wang2011svt} & -- & -- & 257 & 647 & -- \\
        \rownumber & ICDAR2003~\cite{lucas2003icdar} & -- & -- & 1156 & 1110 & -- \\
        \rownumber & ICDAR2013~\cite{karatzas2013icdar} & 229 & 233 & 848 & 1095 & 4.2 \\
        \rownumber & ICDAR2015~\cite{karatzas2015icdar} & 1000 & 500 & 4468 & 2077 & 4.4 \\
        \rownumber & SVTP~\cite{phan2013svtp} & -- & -- & -- & 645 & -- \\
        \rownumber & CUTE80~\cite{risnumawan2014cute} & -- & -- & -- & 288 & -- \\
        \rownumber & Total-Text~\cite{chng2019tt} & 1255 & 300 & 9276 & 2215 & 7.4 \\
        \rownumber & MSRA-TD500~\cite{yao2012msra} & 300 & 200 & -- & -- & -- \\
        \rownumber & CTW-1500~\cite{liu2017ctw} & 1000 & 500 & -- & -- & -- \\
        \rownumber & COCO-Text~\cite{veit2016coco} & 18895 & 4416 & 61793 & 13910 & 3.2 \\
        \cmidrule{1-7}
        \rownumber & ICDAR17-MLT~\cite{nayef2017mlt}$^{*\ddagger}$ & 9000 & 9000 & 85094 & n/a & 9.5 \\
        \rownumber & ICDAR17-RCTW~\cite{shi2017rctw}$^*$ & 8034 & 4229 & 47107 & n/a & 5.9\\
        \rownumber & ICDAR19-MLT~\cite{nayef2017mlt}$^*$ & 10000 & 10000 & 89407 & n/a & 8.9\\
        \rownumber & ICDAR19-ArT~\cite{chng2019art}$^*$ & 5603 & 4563 & 50042 & n/a & 8.9 \\
        \rownumber & ICDAR19-LSVT~\cite{sun2019lsvt}$^*$ & 30000 & 20000 & 243385 & n/a & 9.1\\
        \cmidrule{1-7}
        \bf \rownumber & \bf \datasetName (ours)$^\ddagger$ & \bf 24902 & \bf 3232 & \bf 822572 & \bf 80497 & 32.1 \\
        \bottomrule
    \end{tabular}
    \caption{\textbf{\datasetName vs other datasets.} We only count non-empty words and images with at least one instance). For non end-to-end and test-only datasets, unavailable fields are left blank.~$^*$~ $\Rightarrow$ multi-lingual datasets with no test labels and small English annotations, $^\dagger$~$\Rightarrow$ synthetic datasets. $^\ddagger$~$\Rightarrow$ val set counted in the train set.}
    \label{tab:datasets}
\end{table}
\begin{table}[t]
\footnotesize
\setlength{\tabcolsep}{3pt}
\centering
    \begin{tabular}{@{}lrrrr@{}}
        \toprule
        \bf Count Type & \bf Train & \bf Val & \bf Test & \bf Total \\
        \cmidrule[\heavyrulewidth]{1-5}
        Images & 21749 & 3153 & 3232 & 28134 \\
        \bf Labeled instances & \bf 1052001 & \bf 150338 & \bf 117725 & \bf 1320064 \\
        \cmidrule{1-5}
        Empty words & 337815 & 41952 & 37228 & 416995 \\
        \bf Non-empty words & \bf 714186 & \bf 108386 & \bf 80497 & \bf 903069 \\
        \cmidrule{1-5}
        Non-alphanumeric & 102744 & 15595 & 11596 & 129935 \\
        Less than 3 chars & 197100 & 28726 & 24643 & 250469 \\
        \bf Alphanumeric \& 3+ chars & \bf 414342 & \bf 64065 & \bf 44258 & \bf 522665 \\
        \cmidrule{1-5}
        Rotated (degree $>$ 20) & 118547 & 18548 & 13102 & 150197 \\
        Curved (points $>$ 4) & 14368 & 3099 & 1843 & 19310 \\
        \bottomrule
    \end{tabular}
    \caption{\textbf{\datasetName statistics.} Details on instance types}
    \label{tab:stats}
\end{table}

Table~\ref{tab:datasets} shows the statistics of \datasetName compared to other public datasets. \datasetName has more images than most existing public datasets except for ICDAR19-LSVT~\cite{sun2019lsvt}, a bilingual dataset focusing more on street view images in Chinese. \datasetName has much larger number of annotated words than any existing public datasets, with at least 3x more words than ICDAR19-LSVT and 10x more than the rest. ICDAR19-LSVT contains only 44K words in English, while \datasetName contains predominantly English, with 20x English words. As a result, \datasetName contains on average 32.1 words per images, 3x more than any existing datasets, making it a great source for both word-level text recognition task and image-level text spotting task in text heavy scenes.

Table~\ref{tab:stats} offers more detailed statistics into \datasetName. There are a total of 1.32M labeled instances in \datasetName if including empty word annotations where the word box or polygon is labeled but the text is not transcribed (due to illegibility or language). If we remove words that are non-alphanumeric (e.g. symbols) or have fewer than 3 characters (a standard in some datasets), \datasetName still contains 523k words. Among these, 150k are rotated words ($>20^{\circ}$ angle) and 19.3k are curved words (more than 4 points used to draw the polygon), almost twice the total words in Total-Text~\cite{chng2019tt}, a dataset focusing on curved text.

\section{OCR Experiments}
In this section, we evaluate the \datasetName dataset and the challenge it presents, then exhibit its usefulness and empirically show how it can be used for both training superior models and surpassing existing baselines on other text-recognition benchmark. We demonstrate this through three types of experiments: (i) cross-dataset empirical analysis, (ii) achieving state-of-the-art on public benchmarks, and (iii) evaluating state-of-the-art methods on \datasetName. Please refer to supplementary material for implementation details.

\begin{table*}[t]
    \centering
    \footnotesize
    \setcounter{magicrownumbers}{0}
    \setlength{\tabcolsep}{3.5pt}
    \begin{tabular}{@{}llclrrrrrrrrr@{}}
        \toprule
        \multirow{2}[3]{*}{\bf \#} & \multirow{2}[3]{*}{\bf Method} & \multirow{2}[3]{*}{\bf PW} & \multirow{2}[3]{*}{\bf Train Dataset} & \multicolumn{9}{c}{\bf Test Dataset (Word accuracy)} \\
        \cmidrule{5-13}
        & & & & \bf IIIT & \bf SVT & \bf IC03 & \bf IC13 & \bf IC15 & \bf SVTP & \bf CUTE &  \bf COCOText &  \bf \datasetName \\
        \cmidrule[\heavyrulewidth]{1-13}
        \multicolumn{13}{c}{\bf Cross Dataset Experiments} \\
        \midrule
        \rownumber & TPS-ResNet-BiLSTM-Attn~\cite{baek2019STR} & & COCOText & 70.73 & 73.57 & 85.58 & 82.73 & 64.05 & 60.31 & 50.87 & 53.47 & 43.20 \\
        \rownumber & TPS-ResNet-BiLSTM-Attn~\cite{baek2019STR} & & TextOCR  & \bf 80.50 & \bf 82.84 & \bf 92.16 & \bf 91.25 & \bf 80.07 & \bf 76.74 & \bf 70.38 & \bf 64.03 & \bf 65.65 \\
        \midrule
        \multicolumn{13}{c}{\bf  Benchmarking state-of-the-art models and fine-tuning on \datasetName} \\
        \midrule
        \rownumber & CRNN~\cite{shi2017crnn} & $\cmark$ & S90k+ST & 82.63 & 82.07 & 92.96 & 90.55 & 68.85 & 71.01 & 62.37 & 49.00 & 43.07 \\
        \rownumber & CRNN~\cite{shi2017crnn} (ours) & &  S90k+ST+TextOCR & 85.97 & 87.94 & 92.96 & 93.70 & 79.85 & 78.30 & 73.87 & 60.09 & 58.61 \\
        \rownumber & Rosetta~\cite{borisyuk2018}   & $\cmark$ & S90k+ST & 84.00 & 84.08 & 92.39 & 91.13 & 70.29 & 74.73 & 67.60 & 49.66 & 43.16 \\
        \rownumber & Rosetta~\cite{borisyuk2018} (ours)   & &  S90k+ST+TextOCR & 87.50 & 89.80 & 93.77 & 94.52 & 81.28 & 81.40 & 77.35 & 62.61 & 60.85 \\
        \rownumber & STAR-Net~\cite{liu2016starnet} & $\cmark$ & S90k+ST & 86.26 & 86.09 & 94.39 & 91.48 & 75.81 & 76.43 & 72.47 & 53.56 & 48.23 \\
        \rownumber & STAR-Net~\cite{liu2016starnet} (ours) & &  S90k+ST+TextOCR & \bf 90.30 & \bf 92.12 & \bf 94.69 & \bf 95.33 & \bf 86.09 & 83.88 & \bf 83.62 & 67.92 & 66.84 \\
        \rownumber & TPS-ResNet-BiLSTM-Attn~\cite{baek2019STR}  & $\cmark$ & S90k+ST & 87.37 & 87.33 & 95.12 & 93.00 & 78.24 & 80.16 & 74.22 & 56.34 & 50.37 \\
        \rownumber & TPS-ResNet-BiLSTM-Attn~\cite{baek2019STR} (ours)  & &  S90k+ST+TextOCR & 86.70 & 91.50 & 94.23 & 94.63 & 85.15 & \bf 84.19 & 79.44 & \bf 69.15 & \bf 69.49 \\
        \bottomrule
    \end{tabular}
    \caption{\textbf{Text recognition experiments on the \datasetName dataset.} \textbf{PW} means the model uses public available weights. \textbf{S90k} and \textbf{ST} refer to the Synth90k~\cite{jaderberg2014synth90k} and SynthText~\cite{gupta2016synthtext} datasets respectively. Row \#1-2 show the cross-dataset comparison between COCOText~\cite{veit2016coco} and \datasetName. Row \#3-10 show the experiments on state-of-the-art methods that including TextOCR in training can improve their word accuracy on most public benchmarks, as well as their word accuracy on TextOCR test set.}
    \label{tab:ocr_rec}
\end{table*}

\newcommand{\aman}[1]{\textcolor{red}{[AS: #1]}}
\begin{table*}[t]
    \centering
    \setcounter{magicrownumbers}{0}
    \footnotesize
    \setlength{\tabcolsep}{4.5pt}
    \begin{tabular}{llcccccrrrr}
        \cmidrule[\heavyrulewidth]{1-11}
        \multirow{2}[3]{*}{\bf \#} & \multirow{2}[3]{*}{\bf Method} & \multirow{2}[3]{*}{\bf Official} & \multicolumn{4}{c}{\bf Train Dataset} & \multicolumn{4}{c}{\bf Test Dataset (F-measure)} \\
        \cmidrule(lr){4-7} \cmidrule(lr){8-11}
        & &  & \bf SynthText & \bf Public & \bf COCOText & \bf \datasetName & \bf TT (None) & \bf TT (Full) & \bf COCOText & \bf \datasetName \\
        \cmidrule[\heavyrulewidth]{1-11}
        
        \multicolumn{11}{c}{\bf Cross Dataset Experiments} \\
        \midrule

        \rownumber & Mask TextSpotter v3~\cite{liao2020maskv3}  &          & &          & $\cmark$ & & 54.2 & 65.6 & 52.2 & 32.5 \\
        \rownumber & Mask TextSpotter v3~\cite{liao2020maskv3}  &          & &          & & $\cmark$ & \bf 64.8 & \bf 74.1 & \bf 52.4 & \bf 45.8 \\
        \midrule
        \multicolumn{11}{c}{\bf Benchmarking state-of-the-art models and fine-tuning on \datasetName} \\
        \midrule
        \rownumber & Qin et al. Inc-Res~\cite{qin2019e2e}   & $\cmark$ & $\cmark$ & $\cmark$ & $\cmark$ & & 63.9 & --    & --    & -- \\
        \rownumber & Mask TextSpotter v2~\cite{liao2019maskv2}  & $\cmark$ & $\cmark$ & $\cmark$ &          & & 65.3 & 77.4 & 47.6 & -- \\
        \rownumber & Boundary TextSpotter~\cite{wang2020boundary} & $\cmark$ & $\cmark$ & $\cmark$ &          & & 65.0 & 76.1 & 41.3 & -- \\
        \rownumber & ABCNet~\cite{liu2020abcnet} & $\cmark$ & $\cmark$ & $\cmark$ & $\cmark$ & & 64.2 & 75.7 & --    & 30.5 \\
        \rownumber & Mask TextSpotter v3~\cite{liao2020maskv3}  & $\cmark$ & $\cmark$ & $\cmark$ &          & & 71.2 & 78.4 & 46.1 & 34.9 \\
        \rownumber & Mask TextSpotter v3~\cite{liao2020maskv3} (ours) &    & $\cmark$ & $\cmark$ & & $\cmark$ & \bf 74.5 & \bf 81.6 & \bf 57.9 & \bf 50.8 \\
        \bottomrule
    \end{tabular}
    \caption{\textbf{End-to-end recognition experiments on the \datasetName dataset.} \textbf{Official} means either using official weights (for testing on TextOCR) or offical reported results (other test data). \textbf{Public} refers to the model is trained with public real datasets~\cite{karatzas2013icdar, karatzas2015icdar, chng2019tt, nayef2017mlt} other than COCOText~\cite{veit2016coco} or \datasetName. \textbf{TT} is short for Total Text~\cite{chng2019tt}. Row \#1-2 show the cross-dataset comparison between COCOText and \datasetName. Row \#3-7 show results of state-of-the-art methods, where TextOCR tests are obtained with official weights. Row \#8 show improvements after fine-tuning with TextOCR train data.}
    \label{tab:ocr_e2e}
\end{table*}

\subsection{Cross-dataset empirical analysis}
COCOText~\cite{veit2016coco}, one of the largest fully-annotated English OCR dataset, has images from COCO which were originally collected for object detection purpose, resulting in sparse text occurrences.

As this setting is different from usual OCR applications and benchmarks, COCOText is not ideal to train or test upon. On the other hand, TextVQA dataset (image source of \datasetName) is designed for visual question answering based on text in the image leading to more prominent text, making it also a great OCR data source.

In our text recognition experiments, shown in Table~\ref{tab:ocr_rec}, we use the TPS-ResNet-BiLSTM-Attn model~\cite{baek2019STR}, and train it on COCOText (row \#1) and \datasetName (row \#2) separately from scratch for 100k iterations keeping all other settings the same. We evaluate and compare the results on \datasetName, COCOText and other common text recognition benchmarks. The model trained on \datasetName is 22.45\% better than COCOText on the \datasetName test set, and 10.56\% better even on the COCOText's test set. On the other benchmarks, \datasetName-trained model is consistently better with 10\% or more gap. The superior performance can be attributed to the sheer amount of difference in number of words in these datasets compared to \datasetName.
Note that training on COCOText alone (w/o synthetic data) only achieves 64.05\% word accuracy on ICDAR 2015~\cite{karatzas2015icdar}, signaling it is not a good representative of oriented scene text. Comparatively, training on \datasetName alone can achieve near state-of-the-art performance of 80.07\%. Besides its large scale, we also show \datasetName has good quality compared to previous datasets, by experiments on the same number of instances as ICDAR15 and COCO-Text. Results show \datasetName is 2.5\% better than ICDAR15 on average in recognition benchmarks, and 0.3\% better than COCO-Text, thanks to its good quality and diversity. Please refer to supplementary experiment details.

Table~\ref{tab:ocr_e2e} shows results on end-to-end recognition evaluating \datasetName's usefulness on the image-level task. We use the latest Mask TextSpotter (MTS) V3~\cite{liao2020maskv3}~\footnote{\href{https://github.com/MhLiao/MaskTextSpotterV3}{https://github.com/MhLiao/MaskTextSpotterV3}} and train it from scratch (with ResNet50 trunk pretrained on ImageNet) on COCOText (row \#1) and \datasetName (row \#2) separately. We can see model fine-tuned on \datasetName again has a 0.2\% lead over COCOText on its own test set, and 10\%+ lead on the \datasetName and Total-Text test sets. This demonstrates the advantage of using \datasetName as a training data as it is more generalizable on other datasets. Since the number of images in \datasetName is comparable to COCO-Text (21749 vs 18895), this result is another evidence of \datasetName's good quality.

\subsection{State-of-the-art on public benchmarks}

In this section, using text recognition and end-to-end experiments again, we demonstrate that \datasetName is complementary to existing datasets, and training with it  can improve model accuracy significantly on existing public benchmarks, and even outperform state-of-the-art.

For text recognition, we evaluate the state-of-the-art models based upon Baek et al.~\cite{baek2019STR}~\footnote{\href{https://github.com/clovaai/deep-text-recognition-benchmark}{https://github.com/clovaai/deep-text-recognition-benchmark}}, as well as fine-tune them on \datasetName's train set in Table~\ref{tab:ocr_rec} (rows \#3-10). For each method, fine-tuning on \datasetName brings a significant increase on almost all datasets.  The irregular datasets (\eg ICDAR2015~\cite{karatzas2015icdar}, SVT Perspective~\cite{phan2013svtp} and CUTE80~\cite{risnumawan2014cute}), gain most thanks to the rich diversity in \datasetName.

\begin{figure*}[!ht]
\begin{center}
    \centering
    \includegraphics[width=0.99\textwidth]{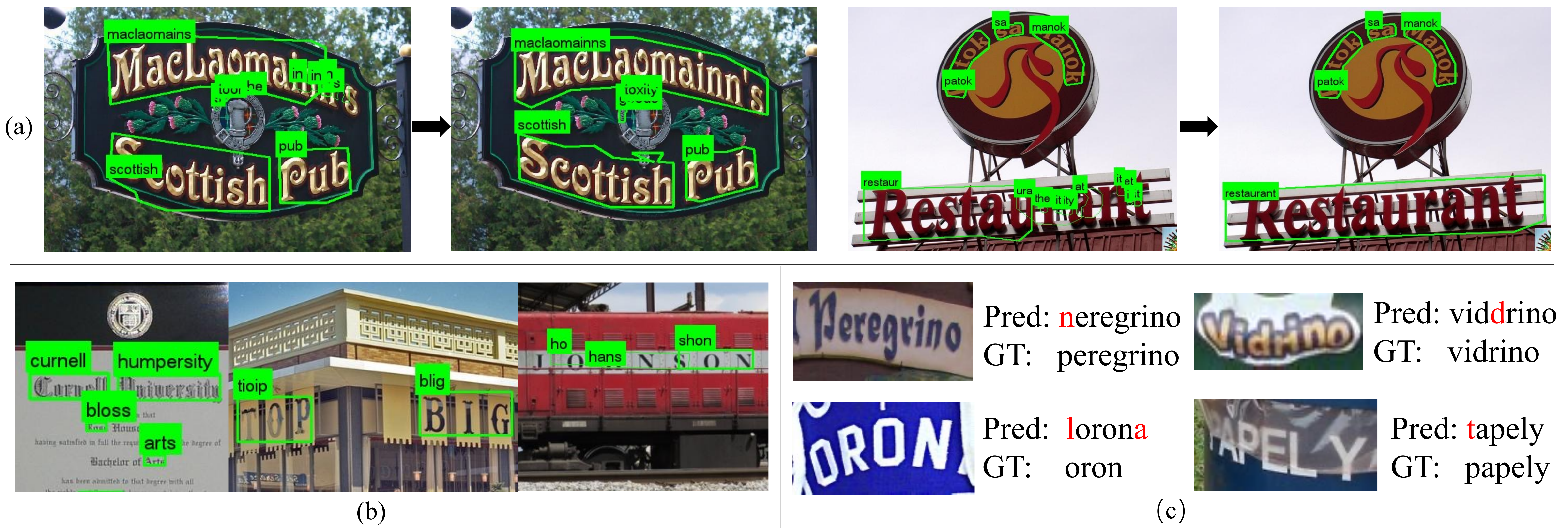}\vspace{-3mm}
\end{center}
   \caption{(a) \textbf{Examples of Mask TextSpotter V3}~\cite{liao2020maskv3} improvement on Total Text after fine-tuning on \datasetName compared to official weights; (b) \textbf{Failure cases by MTS V3} on \datasetName test set; (c) \textbf{Failure cases by Baek et al.}~\cite{baek2019STR} on \datasetName test set}
\label{fig:ocr_results}
\end{figure*}

For end-to-end recognition, we fine-tuned the official weights by Mask TextSpotter V3 on \datasetName and Total Text. Table~\ref{tab:ocr_e2e} (rows \#3-8) again shows that adding \datasetName can further improve the F-measure on Total Text test set by 3.3\% and 3.2\% with none and weak lexicon respectively. Figure~\ref{fig:ocr_results}(a) shows qualitative examples of the results.

\subsection{The challenges of \datasetName}
Following others~\cite{eidinger2014age}, we show the challenges of \datasetName, by evaluating pre-trained and \datasetName fine-tuned state-of-the-art methods on \datasetName test set. The end-to-end recognition results on \datasetName are evaluated in the same protocol as described in~\cite{liao2019maskv2} following ICDAR2015 with support for polygon representation. All experiments were performed with a input short side of 1000 for fair comparison. Note that \datasetName can benefit from higher short sides due to its high resolution.

Table~\ref{tab:ocr_rec}~rows~\#3-10 and Table~\ref{tab:ocr_e2e}~rows~\#3-8 ``\datasetName'' column shows performance of state-of-the-art methods on text and end-to-end recognition tasks, respectively. The results demonstrate \datasetName's challenge; even after fine-tuning with its own large train set of 21k images, the numbers are still much lower than other popular OCR datasets \cite{karatzas2015icdar,chng2019tt}, indicating a difficult task with a large room for improvement.

\section{TextVQA and TextCaps Experiments}

To evaluate the effectiveness and quality of \datasetName for downstream tasks, we calculate various heuristics and conduct experiments on TextVQA and TextCaps dataset using PixelM4C with \datasetName trained OCR module.

\subsection{Upper Bounds and Heuristics}
\label{subsec:textvqa_heuristics}

First, we set new precedents for the TextVQA dataset in Table~\ref{table:textvqa_heuristics} by recalculating the OCR-based upper bounds (UB) and heuristics for its val set presented in \cite{singh2019towards} using Rosetta \cite{borisyuk2018rosetta} OCR-en namespace, OCR tokens from \datasetName trained MTS v3 \cite{liao2020maskv3, multiplexer}, and the annotated text present in \datasetName. 

\begin{table}[h]
\setcounter{magicrownumbers}{0}
\footnotesize

\setlength{\tabcolsep}{10pt}
\centering
    \begin{tabular}{@{}ll@{}rrr@{}}
        \toprule
        \multirow{2}[3]{*}{\bf \#} & \multirow{2}[3]{*}{\bf Method} & \multicolumn{3}{c}{\bf TextVQA val accuracy(\%)} \\
        \cmidrule(lr){3-5}
        & & \multicolumn{1}{c}{\bf Rosetta} & \bf MTS v3 & \bf \datasetName \\
        \cmidrule[\heavyrulewidth]{1-5}
		\rownumber & Human & 85.01 & 85.01 & 85.01\\
		\rownumber & OCR UB & 44.98 & 53.34 & 66.90\\
		\rownumber & Vocab UB & 59.02 & 59.02 & 59.02\\
		\rownumber & OCR + Vocab UB & 79.72 & 80.64 & 87.22\\
		\rownumber & OCR Biggest & 12.06 & 13.60 & 16.78 \\
		\rownumber & OCR Max & 9.26 & 7.50 & 10.94\\
		\bottomrule
    \end{tabular}
    
    \caption{\textbf{TextVQA heuristics.} Val accuracy for various heuristics compared with  numbers from \cite{singh2019towards}. The comparison shows that \datasetName leads to much higher numbers than the original OCR tokens used in the TextVQA.
    }
    \label{table:textvqa_heuristics}\vspace{-5mm}
\end{table}
\begin{table*}[ht]
\setcounter{magicrownumbers}{0}
\footnotesize

\setlength{\tabcolsep}{3pt}
\centering
    \setcounter{magicrownumbers}{0}
    \begin{tabular}{@{}llllrrrrrr@{}}
        \toprule
        \multirow{2}[3]{*}{\bf \#} & \multirow{2}[3]{*}{\bf Method} & \multirow{2}[3]{*}{\bf OCR Source} & \multirow{2}[3]{*}{\bf OCR Feature} &  \thead{\bf TextVQA} & \multicolumn{5}{c}{\bf TextCaps val metrics}\\
        \cmidrule(lr){5-5}\cmidrule(lr){6-10}
        & & & &  \bf val acc (\%) & \bf B-4 & \bf M & \bf R & \bf S & \bf C \\
        \cmidrule[\heavyrulewidth]{1-10}
		\rownumber & Human & -- & -- & 
		85.01  & 24.40 & 26.10 & 47.00 & 18.80 & 125.50 \\
		\rownumber & M4C / M4C-Captioner & Rosetta (OCR-en) & Object detector fc7 & 
		39.40 & 23.30 & 22.00 & 46.20 & 15.60 & 89.60 \\
		\rownumber & M4C w/ STVQA & Rosetta (OCR-en) & Object detector fc7 & 
		40.55 & -- & -- & -- & -- & -- \\
		\rownumber & PixelM4C / PixelM4C-Captioner & MTS v3 (COCO-Text+TT) & MTS v3 fc7 & 
		37.61 & 23.09 & 21.08 & 45.55 & 14.51 & 81.44\\
		\rownumber & PixelM4C / PixelM4C-Captioner & MTS v3 (COCO-Text+TT) & MTS v3 LH &
		38.24 & 23.05 & 20.88 & 45.45 & 14.23 & 81.55\\
		\rownumber & PixelM4C / PixelM4C-Captioner & MTS v3 (\datasetName-en) & MTS v3 fc7 &  
		39.69 & 23.11 & 21.37 & 45.57 & 14.68 & 84.54\\
		\rownumber & PixelM4C / PixelM4C-Captioner & MTS v3 (\datasetName-en) & MTS v3 LH &  
		40.67 & 23.41 & 21.45 & 45.69 & 14.75 & 86.87\\
		\rownumber & PixelM4C / PixelM4C-Captioner & MTS v3 (\datasetName) & MTS v3 fc7 & 39.64 & 23.01 & 21.27 & 45.65 & 14.58 & 84.99\\
		\rownumber & PixelM4C / PixelM4C-Captioner & MTS v3 (\datasetName) & MTS v3 LH & 
		41.23 & 23.33 & 21.30 & 45.71 & 14.62 & 85.32\\
		\rownumber & PixelM4C w/ STVQA & MTS v3 (\datasetName) & MTS v3 LH &
		42.12 & -- & -- & -- & -- & --\\
		\midrule
		\rownumber & PixelM4C / PixelM4C-Captioner & TextOCR & MTS v3 fc7 (from 8)  & 
		46.28 & 23.76 & 21.86 & 46.38 & 15.14 & 91.44 \\
		\rownumber & PixelM4C / PixelM4C-Captioner & TextOCR & MTS v3  LH (from 9)&  
		46.36 & 24.10 & 21.98 & 46.65 & 15.08 & 91.99\\
		\rownumber & PixelM4C w/ STVQA & TextOCR & MTS v3 LH (from 9)& 
		48.04 & -- & -- & -- & --\\
		
		\bottomrule
	\end{tabular}\\
	B--4 = Bleu4 \cite{bleu}, M = METEOR \cite{banerjee2005meteor}, R = ROUGE\_L \cite{lin2004rouge}, M = METEOR \cite{banerjee2005meteor}, C = CIDEr \cite{vedantam2015cider}
    \caption{\textbf{PixelM4C experiments on TextVQA/TextCaps.} Val accuracy for ablations compared with M4C \cite{hu2020iterative}. We show that OCR tokens and features from \datasetName trained models and directly help TextVQA and TextCaps models significantly.
    }\vspace{-3mm}
    \label{table:textvqa_results}
\end{table*}
The \textbf{human} accuracy (row \#1) \cite{singh2019towards} stays the same at 85.01\%. For UB, unlike \cite{singh2019towards}, inspired by M4C's iterative answer prediction, we calculate the accuracy using multi-word match checking whether the answer can be built using single or multiple token(s) from the source in consideration to cover all possibilities allowing better estimates M4C-like models' UB. \textbf{OCR UB} (row \#2) shows the UB achievable by only using OCR tokens and no vocabulary which is 11\% and 22\% higher for \datasetName compared to MTS v3 and Rosetta justifying the requirement of a better OCR mechanism while suggesting that training OCR systems on \datasetName would be crucial for TextVQA \cite{singh2019towards} and TextCaps  \cite{sidorov2020textcaps}. \textbf{Vocab UB} (row \#3) shows the UB achievable by only using a fixed word vocabulary (M4C 5k vocab).
\textbf{OCR+Vocab UB} (row \#4) is UB achievable using both OCR and vocab inter-changably wherever suitable for prediction. For \datasetName, this surpasses the human accuracy indicating \datasetName's high quality and the downstream benefits of improved OCR models. \textbf{OCR Biggest} and \textbf{OCR Max} (row \#5 and \#6) show the UB obtained by choosing biggest OCR box and the most occurring word in the scene text as an answer respectively advocating TextVQA's difficulity, \datasetName's quality and improvement room in current OCR systems. 

\subsection{Improving the state-of-the art}
\label{subsec:textvqa_results}
Given the positive results in Section~\ref{subsec:textvqa_heuristics}, we naturally expect that \datasetName will help with downstream tasks as well, as we know from literature \cite{Gao2020StructuredMA,Kant2020SpatiallyAM} that OCR is indeed an important aspect. Using \datasetName annotations directly will allow us to evaluate the reasoning capabilities or shortcomings of the TextVQA/TextCaps models in isolation from OCR inconsistencies. Furthermore, this also makes it possible to train an end-to-end model that can take images directly as an input, extract OCR tokens from them and then jointly reason over the object features, OCR and input text with a possibility of backpropagating via the recognition model. 

We propose an end-to-end model, PixelM4C shown in Figure~\ref{fig:pipeline}, that works directly on the images allowing us to test our hypotheses. Specifically, we connect the Mask TextSpotter (MTS) v3 trained on \datasetName with M4C. We extract the OCR tokens and features on-the-fly from MTS v3 and pass them to M4C model allowing more fine-grained control on which features to extract and which specific parts to use based on the downstream task and model. We achieve new state-of-the-art on TextVQA using PixelM4C which allows easy testing of our various hypotheses. 

\textbf{Training. }  We train PixelM4C and PixelM4C-Captioner (similar to M4C-Captioner) in an end-to-end fashion by extracting OCR tokens from MTS v3 in real time. We use same hyper-parameters and 5k vocabulary as used by M4C \cite{hu2020iterative} but we set batch size to 16 given that model is slow and hard to train on larger batch sizes. We train with Adam \cite{Kingma2015AdamAM} optimizer with 1e-4 learning rate, a step schedule and a linear warmup of 1k iterations. We train PixelM4C and PixelM4C-Captioner for 24k and 12k iterations. We decrease the learning rate to 1/10th at 14k and 19k for PixelM4C and 10k and 11k for PixelM4C-Captioner. We freeze MTS v3 during training as our empirical results suggested that fine-tuning predictor heads hurt TextVQA accuracy. We hypothesize that this happens because MTS v3 is trained using character-level losses while M4C is trained using word-level losses. Unlike M4C, we conduct ablations on using $>$ 50 tokens given high word density in \datasetName. We train PixelM4C in a distributed fashion on 16 Nvidia Volta V100-SXM2-32GB GPUs using PyTorch based MMF framework \cite{paszke2019pytorch,singh2020mmf}..

\textbf{Experiments and Results. } We compare PixelM4C with M4C for TextVQA and PixelM4C-Captioner with M4C-Captioner for TextCaps. Table~\ref{table:textvqa_results} shows results for various experiments and ablations. First, by an extensive sweep (details in appendix), we confirm that batch size of 16 performs better than batch size of 128 used in \cite{hu2020iterative}. We test PixelM4C and PixelM4C-Captioner with four different OCR sources: MTS v3 trained (i) on COCO-Text and Total-Text (row \#4 and \#5) (ii) on \datasetName but using alphanumeric English only vocabulary (row \#6 and \#7) (iii) on \datasetName using 240 characters Latin vocabulary (row \#8, \#9, and \#10), (iv) using \datasetName annotations directly as the OCR source (row \#11, \#12, and \#13), extracting features using annotation boxes as the proposals from (iii). Enabled by our end-to-end PixelM4C model, we revisit the choice of OCR feature in \cite{hu2020iterative} and try other features from MTS v3. We found that using last hidden state from prediction decoder (``MTS v3 LH'' in Table~\ref{table:textvqa_results}) for $<\mathrm{EOS}>$ token as the OCR representation improves performance. Finally, we add ST-VQA \cite{biten2019scene} as extra annotation data following M4C \cite{hu2020iterative} (row \#10 and \#13).

Based on our ablations (see appendix), we use 200 tokens instead of 50 in all experiments. MTS v3's $\mathrm{fc7}$ features as OCR representation boost accuracy when compared to Visual Genome pretrained FRCNN \cite{Ren2015FasterRT,krishna2017visual} ones (row \#2 vs \#8). Further, we achieve state-of-the-art performance using prediction decoder's last hidden state (row \#9 vs \#3) when compared to $\mathrm{fc7}$ (row \#4 vs \#5, \#6 vs \#7, \#8 vs \#9, and \#11 vs \#12) suggesting that MTS v3 representations including decoder's hidden state contain more relevant information for TextVQA task. Comparing row \#5 with \#7 and \#9, we observe that \datasetName trained models provide better OCR tokens for TextVQA compared to COCO-Text+TT trained ones. Finally, adding STVQA as additional data boosts the performance to 42.12\% setting new state-of-the-art over M4C. In TextCaps, unfortunately, we don't see significant improvement in metrics except B4 using \datasetName trained model's OCR tokens testifying TextCaps's complexity.

Using \datasetName directly as the OCR source gives a significant boost in TextVQA accuracy (6\%) and TextCaps metrics (3\%) signaling that apart from the gap in reasoning capabilities, there is still a room for improvement in OCR capabilities of the OCR module (MTS v3)\footnote{We don't claim this as state-of-the-art because it would be non-ideal for community to train directly on TextOCR except for understanding reasoning capabilities in isolation from OCR systems.}. 
\vspace{-0.25em}
\section{Conclusion}
In this work, we introduced the large arbitrary scene text recognition dataset, \datasetName, collected on TextVQA images along with an end-to-end model, PixelM4C, that can perform scene-text reasoning directly on images by incorporating text recognition model as a module. Training on \datasetName, provides better text-recognition models which outperforms state-of-the-art on most text-recognition benchmarks. Further, using \datasetName trained text-recognition module in PixelM4C allows us to use different features from it with a possibility of even providing feedback which results in PixelM4C surpassing existing state-of-the-art methods on TextVQA. Through \datasetName dataset and PixelM4C model, we take a step towards bridging the communities of OCR and downstream applications based on OCR and hope that research from community will advance both fields at the same time as there is a large room for improvement as evident from TextVQA results from training directly on \datasetName.

{\small
\bibliographystyle{ieee_fullname}
\bibliography{textocr}
}

\input{supp}

\end{document}

%% file: supp.tex
\appendix

\counterwithin{figure}{section}
\counterwithin{table}{section}
\counterwithin{equation}{section}

\twocolumn[{
\begin{center}
\Large 
\textbf{TextOCR: Towards large-scale end-to-end reasoning \\ for arbitrary-shaped scene text}
\\
(Supplementary Material)
\par
\end{center}
\vspace{2em}
}]
\counterwithin{figure}{section}
\counterwithin{table}{section}
\counterwithin{equation}{section}
\newcommand{\modelNameWithoutSpace}{PixelM4C}
\newcommand{\modelName}{\modelNameWithoutSpace\xspace}

\section{Annotation UI Details}

\begin{figure}
    \centering
    \includegraphics[width=\linewidth]{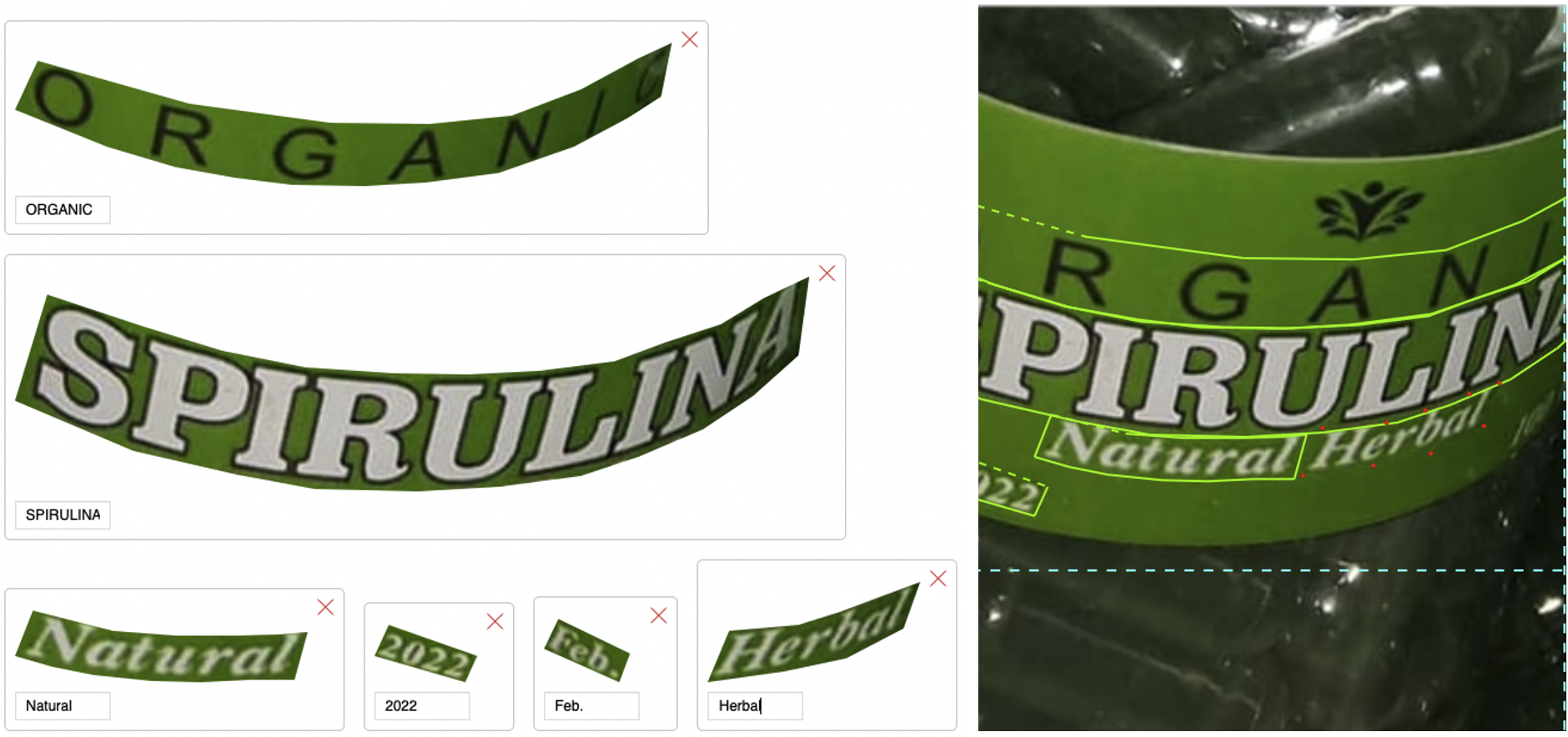}
    \caption{\textbf{Annotation UI} used for TextOCR. UI allows annotating arbitrary shaped text as polygons.}
    \label{fig:ui}
\end{figure}

Figure~\ref{fig:ui} shows the annotation UI that we used for the ground truth labeling of TextOCR. The annotators are able to draw any number of points to form a polygon around arbitrary-shaped word (although they are instructed to draw a quadrilateral whenever appropriate). Each polygon is displayed in a way that the edge between the first and second points is shown differently in a dotted line, to validate that the first point is at the top-left corner of the text, and the points are in clockwise order. Each polygon is then cropped out and displayed on the left screen, where annotators can transcribe the word in the polygon. The UI also has other standard functions such as zoom in/out, panning, delete polygon, and start over. Annotators are also able to re-annotate individual words within an image without needing to start over on the image by clicking `x' on the cropped word. Annotated words are case sensitive. Figure~\ref{fig:stitched} contains more examples of annotated samples. 

\section{Dataset Instance Location Heatmap}
\begin{figure}[t]
\begin{center}
\begin{subfigure}[b]{0.24\linewidth}
\includegraphics[width=\linewidth]{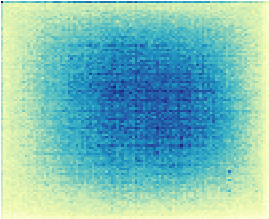}
\caption{TextOCR}
\end{subfigure}
\begin{subfigure}[b]{0.24\linewidth}
\includegraphics[width=\linewidth]{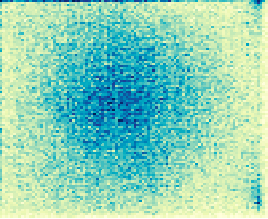}
\caption{COCOText}
\end{subfigure}
\begin{subfigure}[b]{0.24\linewidth}
\includegraphics[width=\linewidth]{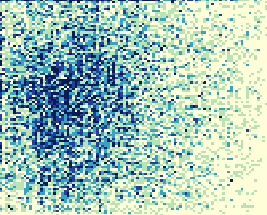}
\caption{ICDAR15}
\end{subfigure}
\begin{subfigure}[b]{0.24\linewidth}
\includegraphics[width=\linewidth]{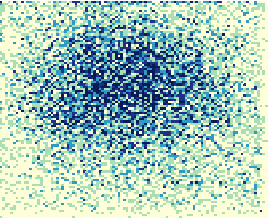}
\caption{TotalText}
\end{subfigure}
\vspace{-10pt}
\end{center}
   \caption{\textbf{Word location heatmap} comparison. TextOCR has more uniform and dense distribution of text instances compared to other datasets.}
\label{fig:heatmap_supp}
\end{figure}

\begin{figure*}
    \centering
    \includegraphics[width=\linewidth]{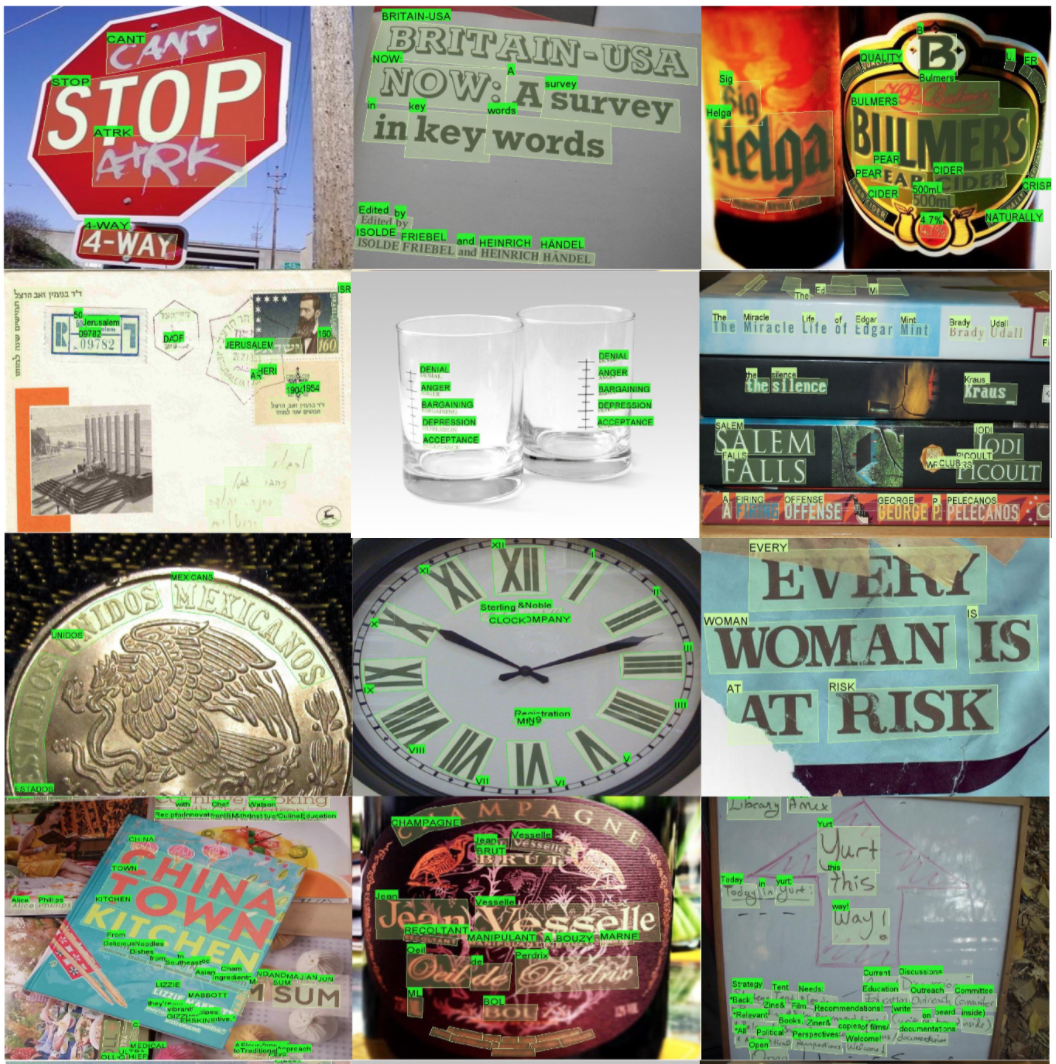}
    \caption{\textbf{More TextOCR annotation samples}}
    \label{fig:stitched}
\end{figure*}

Figure~\ref{fig:heatmap_supp} expands Fig.~\ref{fig:heatmap} in main paper to compare the instance locations of TextOCR, COCO-Text, ICDAR15 and TotalText, an shows TextOCR is more uniformly annotated and distributed and existing datasets.

\section{OCR Model Implementation Details}
We experimented with two types of OCR models in this work, text recognition, and end-to-end recognition.

We use the implementation by Baek et al.~\cite{baek2019STR}~\footnote{\href{https://github.com/clovaai/deep-text-recognition-benchmark}{https://github.com/clovaai/deep-text-recognition-benchmark}} for text recognition task. We experimented with 4 models, including CRNN~\cite{shi2017crnn} (None-VGG-BiLSTM-CTC in~\cite{baek2019STR}), Rosetta~\cite{borisyuk2018} (None-ResNet-None-CTC in~\cite{baek2019STR}), STAR-Net~\cite{liu2016starnet} (TPS-ResNet-BiLSTM-CTC in~\cite{baek2019STR}), and the TPS-ResNet-BiLSTM-Attn model proposed in~\cite{baek2019STR}. For training hyper-parameters, we follow the same settings as in~\cite{baek2019STR} to use AdaDelta optimizer with decay rate of 0.95. The batch size is set to 192, and gradient clipping is applied at a magnitude 5. For the cross-dataset experiments where we are training models from scratch, we train for a total of 200K iterations. For the rest of experiments that fine-tune pretrained models on TextOCR train set, we train for 100K iterations using 4 Tesla Volta V100-SXM2-32GB GPUs. In evaluation, we measure the word accuracy by counting the rate of perfectly predicted words.

For the end-to-end recognition, we use the official implementation of Mask TextSpotter (MTS) V3 by Liao et al.~\cite{liao2020maskv3}~\footnote{\href{https://github.com/MhLiao/MaskTextSpotterV3}{https://github.com/MhLiao/MaskTextSpotterV3}}. We use SGD with momentum of 0.9 and weight decay of 0.0001 for training. The initial learning rate is set to 0.001, and divided by 10 every 100K iterations, for a total of 300K iterations. The batch size is set to 8 and rotation augmentation is performed by randomly rotating input image with an angle between $-$90$\degree$ and 90$\degree$. We also perform multi-scale training that resizes the short side of input image randomly to one of (800, 1000, 1200, 1400). We train our models using 8 Tesla Volta V100-SXM2-32GB GPUs in a distributed fashion using PyTorch \cite{paszke2019pytorch}. During evaluation, we measure with the same protocol as described in~\cite{liao2019maskv2} that follows ICDAR2015 with support for polygon representation, and the short side of input images resized to 1000.

\section{Experiments on same number of instances}
To demonstrate that besides the large scale, TextOCR also has good quality compared to previous datasets, we experimented with the same number of instances as ICDAR15~\cite{karatzas2015icdar} and COCO-Text~\cite{veit2016coco}. We randomly sampled 4055 and 38839 word images from TextOCR for ICDAR15 and COCO-Text, respectively. All experiments fine-tune TPS-ResNet-BiLSTM-Attn~\cite{baek2019STR} from a base pretrained on Synth90k+SynthText, same as paper. As shown in Table~\ref{tab:same_word}, TextOCR-4055 outperforms ICDAR15 on all standard recognition benchmarks except ICDAR15 itself, proving TextOCR provides more diversity and generalizes better to other test sets than ICDAR15, which focuses on incidental scene text. TextOCR-38839 outperforms COCO-Text on 5 out of 7 benchmarks, indicating its superior quality and generalization.

\begin{table}[h]
    \centering
    \footnotesize
    \setcounter{magicrownumbers}{0}
    \setlength{\tabcolsep}{2.8pt}
    \begin{tabular}{@{}llrrrrrrrr@{}}
        \toprule
        \bf \# & \bf Train Dataset & \bf IIIT & \bf SVT & \bf IC03 & \bf IC13 & \bf IC15 & \bf SVTP & \bf CUTE \\
        \midrule
        \rownumber & ICDAR15 & 83.87 & 85.94 & 93.20 & 91.72 & \bf 79.46 & 78.61 & 65.16 \\
        \rownumber & TextOCR-4055 & \bf 87.27 & \bf 88.10 & \bf 94.93 & \bf 93.35 & 78.25 & \bf 80.78 & \bf 72.47 \\
        \midrule
        \rownumber & COCO-Text & 86.07 & 87.79 & \bf 93.66 & 92.77 & 79.79 & 78.61 & \bf 74.91 \\
        \rownumber & TextOCR-38839 & \bf 86.17 & \bf 88.56 & 92.85 & \bf 93.12 & \bf 80.18 & \bf 80.78 & 74.56 \\
        \bottomrule
    \end{tabular}
    \caption{\textbf{Text recognition performance using same number of instances in TextOCR} as in ICDAR15 and COCO-Text. TextOCR achieves better performance indicating its superior quality.}
    \label{tab:same_word}
    \vspace{-10pt}
\end{table}

\section{PixelM4C: Number of OCR tokens}
We conduct a sweep on number of OCR tokens used in \modelName to confirm that more tokens help when the OCR model is trained on TextOCR and the downstream model is using decoder's last hidden state. Table~\ref{tab:num_ocr_tokens}
\begin{table}[h]
\footnotesize
\setcounter{magicrownumbers}{0}
\begin{center}
\begin{tabular}{@{}lllr@{}}
\toprule
\textbf{\#} & \textbf{Experiment} & \textbf{OCR} & \textbf{TextVQA val acc.}\\
\midrule
\rownumber & 50 tokens & MTS v3 (TextOCR-en) & 37.75 \\
\rownumber & 50 tokens & TextOCR & 45.22 \\
\rownumber & 100 tokens & MTS v3 (TextOCR-en) & 39.41 \\
\rownumber & 100 tokens & TextOCR & 46.42 \\
\rownumber & 200 tokens & MTS v3 (TextOCR-en) & 39.41 \\
\rownumber & 200 tokens & TextOCR & 46.12 \\
\rownumber & 200 tokens & MTS v3 (TextOCR-en-LH) & 40.31 \\
\rownumber & 200 tokens & TextOCR-LH & 45.49 \\
\bottomrule
\end{tabular}
\end{center}
\caption{\textbf{Ablation analysis on number of OCR tokens}. The results show that more OCR tokens are better for TextVQA \cite{singh2019towards} when the OCR model is trained on TextOCR.}
\label{tab:num_ocr_tokens}
\end{table}
\section{PixelM4C: Hyper-parameters and ST-VQA}
Table~\ref{tab:supp_hyperparameters} shows various hyper-parameter choices for \modelName and \modelName-Captioner used for training the models on TextVQA \cite{singh2019towards} and TextCaps \cite{sidorov2020textcaps} dataset. We compare the performance of the model on batch size 16 as well as 128 and found batch size 16 reasonably better or equal in performance to batch size 128. For the ease of training the model with less number of GPUs, we stick with batch size 16 for our experiments. 

The confidence threshold for filtering of OCR tokens which works for the best OCR performance doesn't work as it is for \modelName suggesting one more motivation for fine-tuning and adjusting OCR models based on the downstream task. The OCR model (MTS v3) uses a of 0.2 confidence threshold on detection score and 0.8 on recognition score. For \modelName, the no threshold on detection score and 0.2 confidence threshold on recognition score works best which we confirm by a hyper-parameter sweep.
\begin{table}[h]
\footnotesize
\begin{center}
\begin{tabular}{@{}lrr@{}}
\toprule
\textbf{Hyper-parameter} & \textbf{PixelM4C} & \textbf{PixelM4C-Captioner} \\
\midrule
batch size & 16 & 16\\
learning rate & 1e-4 & 1e-4\\
learning schedule & step(14k, 19k) & step(10k, 11k)\\
warmup iterations & 1000 & 1000\\
maximum iteration & 24000 & 12000\\
Adam $\beta_1$ & 0.9 & 0.9\\
Adam $\beta_2$ & 0.999 & 0.99\\
\bottomrule
\end{tabular}
\end{center}
\caption{\textbf{\modelName hyper-parameters.}}
\label{tab:supp_hyperparameters}
\end{table}

For completeness, we also trained PixelM4C with TextOCR trained Latin OCR model on ST-VQA \cite{biten2019scene} train set and test on its validation set created in \cite{hu2020iterative}. We get an accuracy of 38.49\% and 47.89\% ANLS better than that reported in \cite{hu2020iterative} again justifying that TextOCR leads to better downstream models.

\section{Sources of the media used}

\begin{itemize}[nosep,itemsep=1pt,leftmargin=1.5em,labelwidth=*,align=left]
    \item Figure 2 (row 1, column 1), ``The What'' by \href{https://www.flickr.com/people/zimpenfish/}{rjp} licensed CC-BY-2.0.
    \item Figure 2 (row 1, column 2), ``Washington D.C. Tour - African Land Forces Summit - 201005611'' by \href{https://www.flickr.com/people/usarmyafrica/}{US Army Africa} licensed CC-BY-2.0
    \item Figure 2 (row 1, column 3), ``slc camp'' by \href{https://www.flickr.com/people/thefangmonster/}{Noah Sussman} licensed CC-BY-2.0
    \item Figure 2 (row 1, column 4), ``1945'' by \href{https://www.flickr.com/people/homini/}{Homini:)} licensed CC-BY-2.0
    \item Figure 2 (row 2, column 1), ``i\'m watch'' by \href{https://www.flickr.com/people/shisho_1975}{shinji\_w} licensed CC-BY-2.0
    \item Figure 2 (row 2, column 2), ``Cleansui CSP-801'' by \href{https://www.flickr.com/people/othree/}{othree} licensed CC-BY-2.0
    \item Figure 2 (row 2, column 3), ``KA 003'' by \href{https://www.flickr.com/people/kaja_a/}{Kaja Avberšek} licensed CC-BY-2.0
    \item Figure 2 (row 2, column 4), ``Darwin Origin of Species exhibit at Huntington Library'' by \href{https://www.flickr.com/people/mgraessle/}{favouritethings} licensed CC-BY-2.0
    \item Figure 2 (row 3, column 1), ``Greetings from Tallahassee, Florida'' by \href{https://www.flickr.com/people/boston_public_library/}{Boston Public Library} licensed CC-BY-2.0
    \item Figure 2 (row 3, column 2), ``Another design ready for our Print Party. In solidarity with a prisoner led- movement calling for the abolition of solitary confinement. prepping for a big rally and on Tuesday in Sacramento. \#rinitempleton  \#abolishsolitary \#art \#artistactivism \#phss'' by \href{https://www.flickr.com/people/dignidadrebelde/}{dignidadrebelde} licensed CC-BY-2.0
    \item Figure 2 (row 3, column 3), ``Clock -- 1319 F Street NW Washington (DC) July 2013,413654'' by \href{https://www.flickr.com/people/22711505@N05}{Ron Cogswell} licensed CC-BY-2.0
    \item Figure 2 (row 3, column 4), ``Angry Man \#Knock-out'' by \href{https://www.flickr.com/people/philliecasablanca}{Phil Whitehouse} licensed CC-BY-2.0
    \item Figure 4 (b) (left) ``Ross Diploma'' by Ross Housewright
    \item Figure 4 (b) (middle) ``Clark's Big Top Restaurant, 1968'' by \href{https://www.flickr.com/people/seattlemunicipalarchives/}{Seattle Municipal Archives}
    \item Figure 4 (b) (right) ``Locomotive'' by \href{https://www.flickr.com/people/43436168@N00/}{Duane Burdick}
    \item Figure 4 (c) (top left) ``Tienda de souvenirs en santiago'' by \href{https://www.flickr.com/people/compostelavirtua}{compostelavirtual.com}
    \item Figure 4 (c) (top right) ``DSC00062'' by \href{https://www.flickr.com/people/calu777}{Carlos Correa Loyola}
    \item Figure 4 (c) (bottom left) ``I REMEMBERS DAYS OF OLD'' by \href{https://www.flickr.com/people/49889874@N05}{marc falardeau}
    \item Figure 4 (c) (bottom right) ``DSC00062'' by \href{https://www.flickr.com/people/calu777}{Carlos Correa Loyola}   
    \item Figure C1 (row 1, column 1), ``ATRK'' by \href{https://www.flickr.com/people/32492319@N08}{BOMB THE SYSEM} licensed CC-BY-2.0
    \item Figure C1 (row 1, column 2), ``Lost Book'' by \href{https://www.flickr.com/people/bowbrick}{Steve Bowbrick} licensed CC-BY-2.0
    \item Figure C1 (row 1, column 3), ``Big Helga and Bulmers'' by \href{https://www.flickr.com/people/jdennes}{James Dennes} licensed CC-BY-2.0
    \item Figure C1 (row 1, column 4), ``Bull Herzl fifty years to his death (original in Hebrew)'' by \href{https://www.flickr.com/people/zeevveez}{zeevveez} licensed CC-BY-2.0
    \item Figure C1 (row 2, column 1), ``Good Grief Glasses'' by \href{https://www.flickr.com/people/x1brett}{brett jordan} licensed CC-BY-2.0
    \item Figure C1 (row 2, column 2), ``DSC\_0092'' by \href{https://www.flickr.com/people/63742054@N00}{mlwilson1410} licensed CC-BY-2.0
    \item Figure C1 (row 2, column 3), ``cien pesos 1977 4735'' by \href{https://www.flickr.com/people/sirqitous}{Eric Golub} licensed CC-BY-2.0
    \item Figure C1 (row 2, column 4), ``Clock Squircle'' by \href{https://www.flickr.com/people/xurble}{Gareth Simpson} licensed CC-BY-2.0
    \item Figure C1 (row 3, column 1), ``Every Woman Is At Risk'' by \href{https://www.flickr.com/people/p_x_g}{Peter Galvin} licensed CC-BY-2.0
    \item Figure C1 (row 3, column 2), ``Spotted at Kinokunia Books, San Francisco @hollowlegs'' by \href{https://www.flickr.com/people/garysoup}{Gary Stevens} licensed CC-BY-2.0
    \item Figure C1 (row 3, column 3), ``Boozy from Bouzy is our favorite! \#delectable \#wine'' by \href{https://www.flickr.com/people/dalecruse}{Dale Cruse} licensed CC-BY-2.0
    \item Figure C1 (row 3, column 3), ``Yurt Exhibit'' by \href{https://www.flickr.com/people/kirbyurner}{thekirbster} licensed CC-BY-2.0
\end{itemize}